\documentclass[5p,authoryear]{elsarticle} 

\usepackage{framed,multirow}
\usepackage{amssymb}
\usepackage{latexsym}
\usepackage{xcolor}
\usepackage{lineno,hyperref}
\modulolinenumbers[5]
\usepackage{amsmath}
\usepackage{multirow}
\usepackage{color,graphicx}
\usepackage{nicematrix}
\usepackage{arydshln}
\usepackage{hyperref}
\usepackage{url}
\usepackage{footnote}
\usepackage{bm}
\usepackage{cite}
\usepackage{subfigure}
\usepackage{verbatim}
\usepackage{ulem}
\usepackage{graphicx}
\usepackage{pgfplots}
\usepackage{cancel}

\usepackage{booktabs}  
\usepackage{tabularx}
\usepackage{booktabs}
\usepackage{graphicx}
\usepackage{amssymb}
\usepackage{makecell}

\usepackage{enumitem}

\journal{Medical Image Analysis}

\begin{document}

\begin{frontmatter}

\title{Anatomy-Aligned Surface Field Learning for Myocardial Reconstruction from Sparse Short-Axis Cine MRI}

\author[label1,label2]{Xiaohan Yuan} 
\author[label1]{Xuan Yang} 
\author[label1]{Qingya Li}
\author[label2]{Yangang Wang} 
\author[label1]{Lei Li*} 
\ead[url]{lei.li@nus.edu.sg}

\address[label1]{Department of Biomedical Engineering, National University of Singapore, Singapore}
\address[label2]{School of Automation, Southeast University, Nanjing, China}

\begin{abstract}
Patient-specific 4D myocardial reconstruction from cine MRI supports quantitative functional assessment, regional motion analysis, and simulation-based modeling. However, routinely acquired short-axis (SAX) cine MRI is sparsely sampled along the through-plane direction, making dense and anatomically consistent surface reconstruction challenging. In this study, we propose an anatomy-aligned surface learning framework that parameterizes the epicardial and endocardial surfaces on a shared circumferential-longitudinal \textit{UV domain}. This formulation converts irregular 3D reconstruction into structured coordinate-field completion with explicit correspondence across subjects and cardiac phases. Sparse SAX contours are encoded as UV observation fields, coverage-aware sampling improves robustness to incomplete slice coverage, and topology- and distortion-aware learning preserves circumferential continuity and local surface quality. Experiments on three public cine MRI datasets showed that the proposed method consistently outperformed representative mesh-based and implicit reconstruction approaches, achieving overall Chamfer distances of $2.887$~mm on ACDC, $2.641$~mm on M\&Ms, and $2.810$~mm on M\&Ms-2. The reconstructed sequences also preserved ventricular function, with end-diastolic volume and ejection fraction errors of $3.3$~mL and $1.1 \%$, respectively. These results demonstrate that anatomy-aligned UV learning provides an accurate, efficient, and correspondence-aware representation for sparse cine MRI reconstruction and myocardial modeling. The source code will be available at \url{https://github.com/yuan-xiaohan/SAX2MyoSurf}.
\end{abstract}


\end{frontmatter}

\section{Introduction}

\begin{figure}[ht]
    \centering
    \includegraphics[width=0.98\columnwidth]{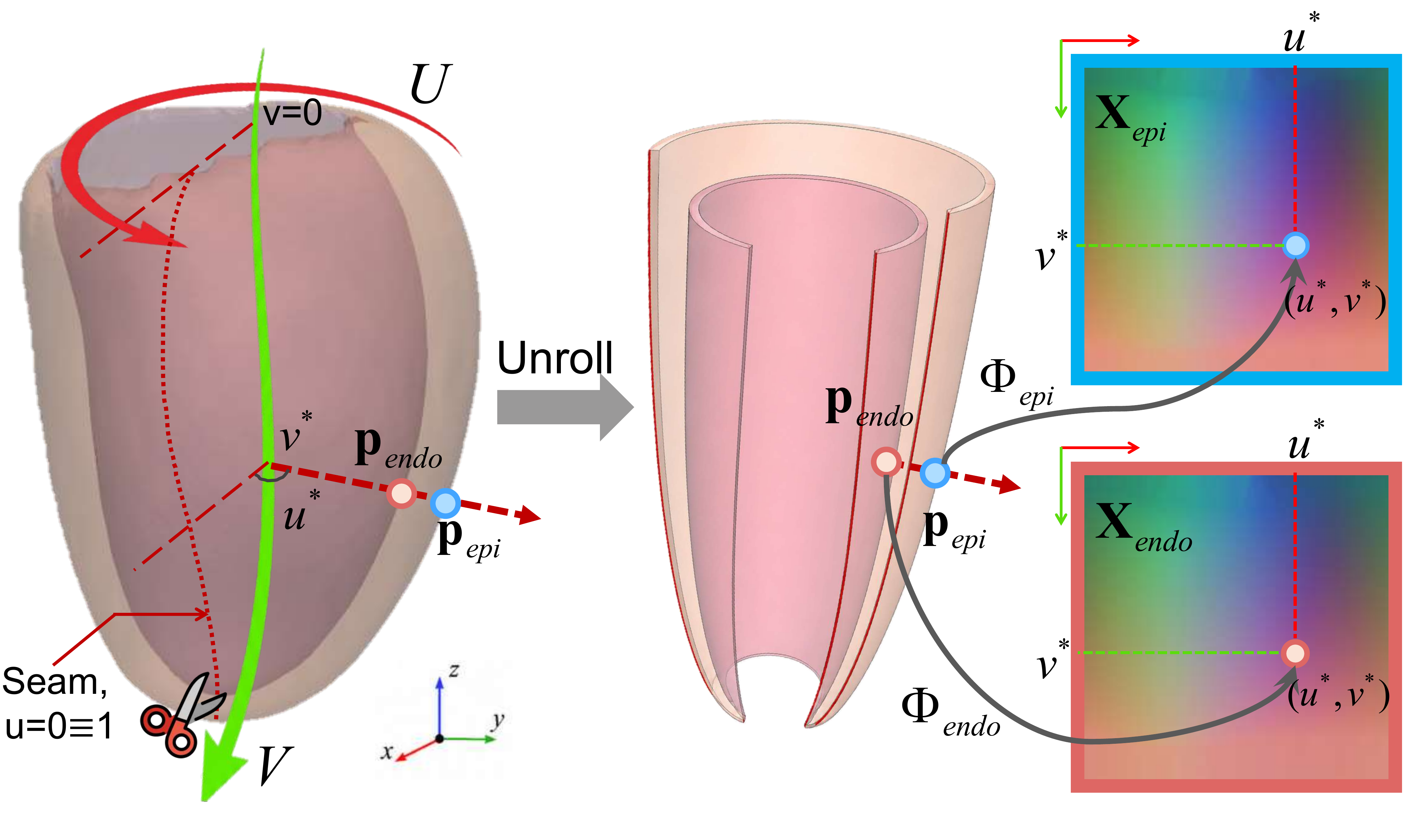}
    \caption{Overview of the anatomy-aligned UV representation for sparse short-axis (SAX) myocardial reconstruction. The epicardial (epi) and endocardial (endo) surfaces are unfolded onto shared circumferential-longitudinal UV domains, where each location stores the corresponding 3D surface coordinates. 
    }
    \label{fig:teaser}
\end{figure}


Patient-specific 4D (3D+t) myocardial mesh reconstruction is fundamental for quantitative cardiovascular analysis \citep{journal/NatureMI/qiao2025,conf/MICCAI/liu2026,journal/RBME/li2024}. 
Compared with image-based measurements alone, surface-based representations provide explicit anatomical geometry and enable downstream analyses such as spatial correspondence, regional functional mapping, and simulation-based modeling. 
In routine clinical imaging, cine MRI is commonly acquired as short-axis (SAX) slices, which provide high in-plane resolution but limited through-plane coverage along the long-axis direction \citep{journal/MedIA/li2023}.
Recovering a geometrically smooth and anatomically plausible myocardial surface from such sparse SAX observations remains challenging because geometric evidence is limited to sparse SAX planes.

Existing learning-based myocardial reconstruction methods can be broadly grouped into explicit mesh deformation and implicit field prediction. 
Mesh-based methods typically deform a predefined template by regressing vertex-wise displacements, often using graph neural networks, mesh convolution, or differentiable projection losses to constrain the reconstructed surface \citep{journal/TMI/meng2022,journal/TMI/meng2023,journal/MedIA/luo2026, journal/MedIA/gaggion2025}. 
These methods have shown promising performance when dense annotations, multi-view images, or strong projection constraints are available. 
However, under sparse SAX supervision, image evidence is only observed at a limited number of slice locations, making it difficult to propagate local contour information coherently across the entire ventricular surface. 
Moreover, vertex-space learning provides consistent topology but lacks an explicit anatomical coordinate system for organizing sparse SAX observations along the myocardial circumferential and longitudinal directions.
Implicit reconstruction methods alleviate some of these limitations by predicting continuous occupancy, signed distance, or coordinate fields in 3D space, thereby improving smoothness and topology preservation \citep{conf/arxiv/chen2024,conf/MICCAI/fu2025,conf/CVPR/sun2022,conf/ICCV/yuan2023,conf/ICCV/ye2023}. 
Nevertheless, most implicit methods still operate in generic Euclidean coordinates and do not explicitly exploit the intrinsic anatomical organization of the myocardium. 
This is suboptimal for the left ventricle (LV), which exhibits a well-defined \textit{circumferential-longitudinal structure with approximately cylindrical topology}. 
Such intrinsic coordinates have been used for regional partitioning, cardiac mechanics, and quantitative cardiac MRI analysis \citep{journal/MedIA/bayer2018,journal/MedIA/schuler2021}. 
In computer graphics and human/object surface modeling, surface parameterization has been widely used to represent complex 3D surfaces on structured 2D domains \citep{conf/CVPR/guler2018,conf/CVPR/zuffi2017, journal/TPAMI/huang2022,journal/TVCG/zhao2023,conf/CVPR/hu2024}. 
However, using such a surface domain to jointly organize sparse SAX observations and dense reconstruction targets remains underexplored.

In this study, we introduce an anatomy-aligned surface learning framework for reconstructing dense 3D myocardial surfaces from sparse SAX cine MRI. 
The proposed method reformulates myocardial reconstruction from sparse slice observations as continuous 3D coordinate field estimation on a shared two-dimensional anatomical surface domain. 
Specifically, we parameterize the LV surface using a circumferential-longitudinal \textit{UV domain}, where the $u$ coordinate follows the circumferential direction and the $v$ coordinate follows the longitudinal base-to-apex direction. 
This representation establishes anatomical correspondence across subjects and converts irregular 3D surface reconstruction into structured surface field learning. 
Sparse SAX contours are further lifted into the UV domain as observation fields, providing spatially coherent geometric conditioning from discontinuous slice measurements. 
To respect the cylindrical topology of the LV, we incorporate explicit periodic modeling along the circumferential direction. 
To our knowledge, this is the first sparse-SAX reconstruction framework to represent both observations and target surfaces as dense fields on the same anatomy-aligned UV domain.
The main contributions of this work are:
\begin{itemize}
    \item We introduce an anatomy-aligned UV representation that reformulates sparse-SAX myocardial reconstruction as structured coordinate-field learning with explicit correspondence across subjects and cardiac phases.

    \item We propose a contour-to-UV encoder that maps discontinuous SAX contours into surface-aligned observation fields, together with coverage-aware sampling for incomplete slice coverage.

    \item We develop a topology- and distortion-aware completion network that combines circumferential wrap convolution with region-focused geometric regularization for coherent epicardial and endocardial reconstruction.

    \item We validate the framework on three public cine MRI datasets, demonstrating accurate and efficient 3D/4D myocardial reconstruction.
\end{itemize}

\section{Related work} \label{sec:related_work}

\subsection{3D/4D Cardiac Reconstruction from Cine MRI}

Existing 3D/4D cardiac reconstruction methods can be broadly categorized into explicit surface modeling, implicit continuous representations, and generative shape modeling. Explicit approaches recover patient-specific anatomy by deforming fixed-topology meshes, templates, or statistical shape models. Representative methods include the MulViMotion series~\citep{journal/TMI/meng2022,journal/TMI/meng2023}, Mesh4D~\citep{conf/MICCAI/qiao2025}, HeartSSM~\citep{journal/arxiv/ma2026}, and atlas-based frameworks~\citep{journal/MedIA/sinclair2022}. Subsequent studies incorporated point-cloud deformation~\citep{journal/JBHI/beetz2024}, CT-derived priors~\citep{conf/STACOM/xu2024}, differentiable slicing~\citep{journal/MedIA/luo2026}, graph-based refinement~\citep{journal/MedIA/gaggion2025}, and statistical shape constraints~\citep{journal/MedIA/joyce2022,journal/MedIA/xia2022}. Although these methods provide explicit and interpretable surfaces, they often depend on multi-view imaging, dense supervision, or strong geometric priors, which limits their applicability to sparse SAX-only acquisitions.

Implicit approaches instead represent cardiac anatomy and motion using continuous fields or deformable domains, including dynamic Gaussian representations~\citep{conf/MICCAI/fu2025}, topology-preserving implicit registration~\citep{conf/CVPR/sun2022}, disentangled shape-motion models~\citep{conf/ICCV/yuan2023}, neural deformation fields~\citep{conf/ICCV/ye2023}, and deformable tetrahedral models~\citep{conf/arxiv/chen2024,conf/arxiv/chen2025}. NIHC~\citep{muffoletto2025neural} further introduced anatomically coherent implicit coordinates based on universal ventricular coordinates. 
In contrast, our formulation explicitly lifts sparse SAX contours into dense surface-aligned observation fields for direct UV-domain completion.
Generative models have also been explored for dynamic shape synthesis, surface completion, and virtual population modeling~\citep{conf/MICCAI/ma2025,journal/NatureMI/qiao2025,conf/MICCAI/zhang2025,conf/MICCAI/yang2026}. Their emphasis, however, is primarily on distribution learning rather than directly reconstructing dense myocardial surfaces from sparse SAX observations while preserving explicit anatomical correspondence across subjects and cardiac phases.

\subsection{Anatomical Coordinates and Canonical Surface Representations}

Cardiac anatomical coordinate systems have been widely used for regional partitioning, functional analysis, and quantitative modeling. 
For example, the AHA 17-segment model provides a standardized regional reference for analyzing myocardial perfusion, motion abnormalities, and scar distribution \citep{journal/TMI/li2024}. 
Universal ventricular coordinates \citep{journal/MedIA/bayer2018} and Cobiveco \citep{journal/MedIA/schuler2021} further construct continuous anatomical coordinate systems for establishing correspondences across cardiac models for cross-subject functional comparison. 
However, these coordinate systems are typically used as post-processing analysis tools, simulation coordinate systems, or registration reference spaces, and have rarely been adopted as the primary learning domain for reconstruction models under sparse observations.
In general computer vision and graphics, canonical UV representations are widely used to unfold deformable 3D surfaces into cross-instance consistent 2D parameter domains, thereby transforming irregular surface learning into image-like representation learning. 
Representative works, such as DensePose \citep{conf/CVPR/guler2018}, canonical animal/human surface modeling \citep{conf/CVPR/zuffi2017}, and SUPPLE \citep{journal/TVCG/zhao2023}, demonstrate the advantages of UV representations in dense correspondence, geometric alignment, and cross-instance structural modeling. 
Nevertheless, these methods are mainly designed for natural images, dense surface observations, or texture-based supervision, and have not been fully explored for sparse medical image-based reconstruction \citep{conf/CVPR/geng2023,conf/CVPR/hu2024}.

\begin{figure*}[h]
\includegraphics[width=1\textwidth]{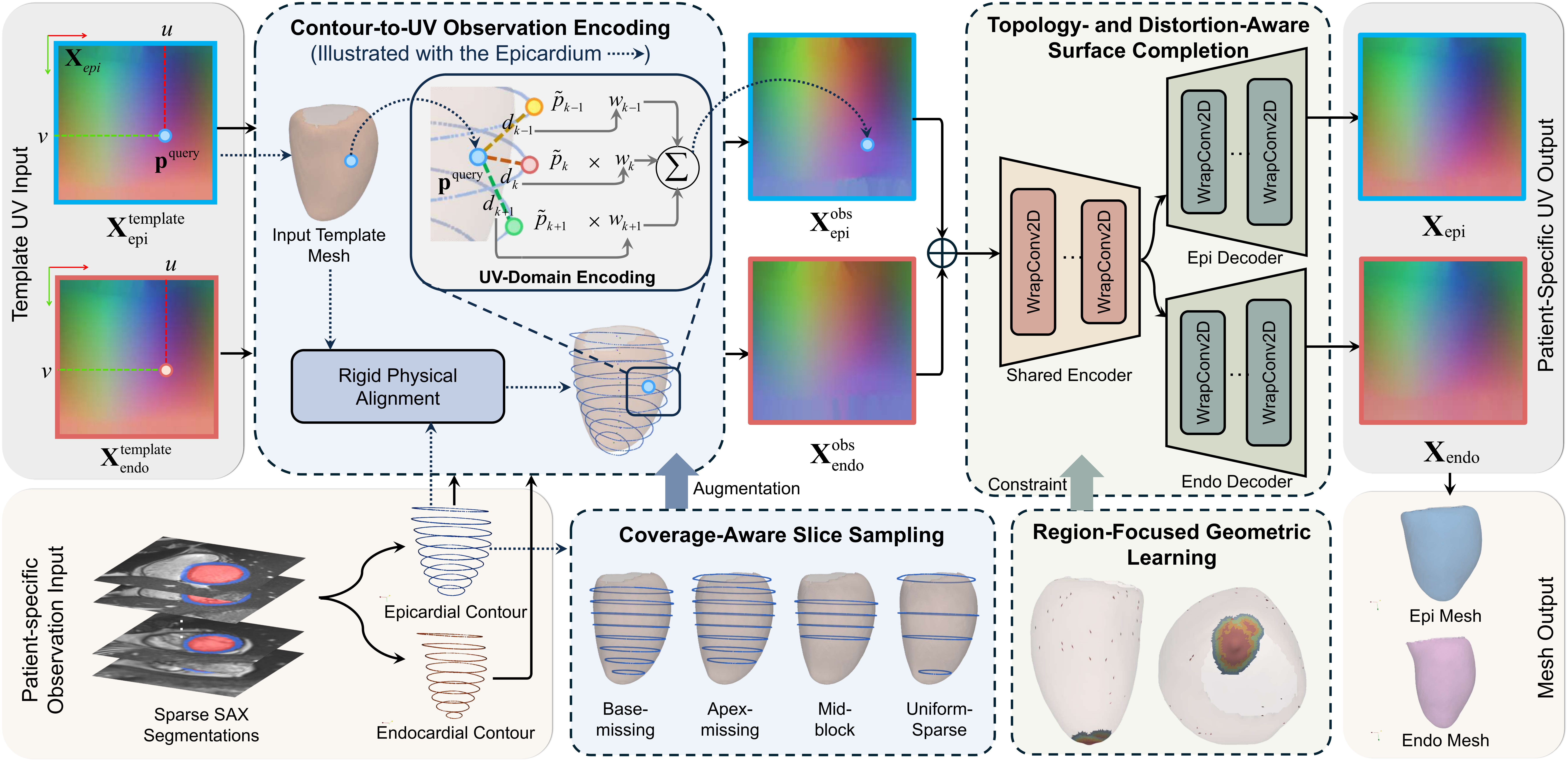}
\vspace{-12pt
}
\caption{Overall pipeline of the proposed anatomy-aligned UV myocardial surface reconstruction framework. Sparse SAX contours are aligned to the canonical template and encoded as epicardial and endocardial UV observation fields, with coverage-aware sampling used during training. A topology- and distortion-aware UV network completes the patient-specific surface fields through circumferential WrapConv2D and region-focused geometric supervision, followed by explicit mesh reconstruction using fixed UV connectivity.}
\label{fig:method:pipeline}
\end{figure*}

\section{Method}

As illustrated in Fig.~\ref{fig:method:pipeline}, the proposed framework reconstructs patient-specific myocardial surfaces from SAX contours through three key components. First, the epicardial and endocardial surfaces are represented as 3D coordinate fields on a shared anatomy-aligned UV domain (Sec.~\ref{sec:A}). Second, sparse SAX contours are transformed into surface-aligned observation fields through anatomical correspondence retrieval and soft cross-slice aggregation, with coverage-aware sampling introduced during training (Sec. \ref{sec:B}). Third, a periodic UV-domain network completes the two surface fields, while topology- and distortion-aware constraints preserve circumferential continuity and suppress local geometric artifacts. The predicted coordinate fields are finally converted into explicit meshes using fixed UV connectivity (Sec. \ref{sec:C}).

\subsection{Anatomy-Aligned UV Surface Representation}
\label{sec:A}

As illustrated in Fig.~\ref{fig:teaser}, we represent the epicardial and endocardial surfaces on a shared circumferential-longitudinal UV domain. A myocardial template \citep{bai2015bi} provides consistent topology and vertex correspondence across subjects and phases. Its UV parameterization is constructed once and shared across all reference and predicted surfaces, such that the same UV location corresponds to the same anatomical position across subjects and cardiac phases. Specifically, we define the canonical domain as
\begin{equation}
\Omega
=
\left\{
(u,v)
\mid
u\in[0,1),\;
v\in[0,1]
\right\},
\label{eq:canonical_domain}
\end{equation}
where $u$ is the periodic circumferential coordinate and $v$ is the
longitudinal coordinate from the base to the apex. For each surface
$s\in\{\mathrm{epi},\mathrm{endo}\}$, the parameterization is defined as
\begin{equation}
\Phi_s:
\mathcal{S}^{\mathrm{template}}_s
\rightarrow
\Omega,
\qquad
\mathbf{x}
\mapsto
\left(
u_s(\mathbf{x}),
v_s(\mathbf{x})
\right).
\label{eq:surface_parameterization}
\end{equation}
The two surfaces share the same centerline, septal reference, and seam
convention, maintaining correspondence between the myocardial
boundaries.

Let $\mathbf{c}(\tau)$ denote the template centerline parameterized by
normalized arc length $\tau\in[0,1]$, from the base to the apex. The
longitudinal coordinate of a surface point $\mathbf{x}$ is determined
by its nearest centerline position:
\begin{equation}
\tau^{*}(\mathbf{x})
=
\underset{\tau\in[0,1]}{\arg\min}\,
\left\|
\mathbf{x}-\mathbf{c}(\tau)
\right\|_2,
\qquad
v_s(\mathbf{x})=\tau^{*}(\mathbf{x}).
\label{eq:longitudinal_coordinate}
\end{equation}
At $\mathbf{c}(\tau^{*})$, the centerline tangent
$\mathbf{t}(\tau^{*})$ defines the local cross-sectional plane. A
septal reference projected onto this plane defines
$\mathbf{e}_1(\tau^{*})$, and
$\mathbf{e}_2(\tau^{*})
=\mathbf{t}(\tau^{*})\times\mathbf{e}_1(\tau^{*})$
completes the local orthonormal basis. The circumferential coordinate is
then
\begin{equation}
u_s(\mathbf{x})
=
\frac{
\left[
\operatorname{atan2}
\left(
\mathbf{r}\cdot\mathbf{e}_2(\tau^{*}),
\mathbf{r}\cdot\mathbf{e}_1(\tau^{*})
\right)
-
\theta_{\mathrm{seam}}
\right]
\bmod 2\pi
}{
2\pi
},
\label{eq:circumferential_coordinate}
\end{equation}
where $\mathbf{r}
=
\mathbf{x}-\mathbf{c}(\tau^{*})$, and $\theta_{\mathrm{seam}}$ fixes the seam at a consistent anatomical
location.
The resulting parameterization represents each surface as a
3D coordinate field:
\begin{equation}
\mathbf{X}_s:
\Omega_s
\rightarrow
\mathbb{R}^{3},
\qquad
(u,v)
\mapsto
\mathbf{X}_s(u,v),
\label{eq:surface_coordinate_field}
\end{equation}
where $\Omega_s\subseteq\Omega$ is the valid region of surface $s$, and $\mathbf{X}_s(u,v)$ stores the corresponding 3D position. We denote the template and patient-specific fields by $\mathbf{X}^{\mathrm{template}}_s$ and $\mathbf{X}^{*}_s$, respectively.
Note that the coordinate fields are discretized on an $H\times W$ grid with a surface-specific validity mask $M_s$. Fixed connectivity over the valid grid locations converts each field into an explicit triangular mesh.
Therefore, irregular 3D myocardial reconstruction is
reformulated as structured coordinate-field learning on a shared
anatomical domain.

\subsection{Sparse SAX Contour-to-UV Observation Modeling}
\label{sec:B}

The canonical UV representation defines the reconstruction target but does not directly organize the sparse SAX contour observations. We therefore first transform the discontinuous contours into observation fields on the same anatomy-aligned domain, and further introduce coverage-aware slice sampling to improve robustness to varying longitudinal coverage.

\subsubsection{Contour-to-UV Observation Encoding}

The SAX contours are first rigidly aligned with the canonical template. Let $\mathcal{C}_{s,k}=\{\mathbf{p}_{s,k,i}\}_{i=1}^{N_{s,k}}$ denote the aligned contour of surface $s\in\{\mathrm{epi},\mathrm{endo}\}$ on slice $k$. For each canonical UV location $(u,v)$, the corresponding template point $\mathbf{p}^{\mathrm{query}}_s=\mathbf{X}^{\mathrm{template}}_s(u,v)$ is used as the anatomical query. Since each UV location has a fixed anatomical meaning, the template query provides a stable anatomical prior for associating sparse contour observations with canonical surface locations.

Before retrieval, $\Psi(\cdot)$ places each contour at its anchor-based centerline position and aligns its normal with the local tangent using shared end-diastolic (ED) parameters. The transformed contour points can then be compared directly with the template query using Euclidean distance. For each slice, we retrieve the transformed contour point closest to the query and select its corresponding point from the rigidly aligned contour:
\begin{equation}
i^{*}_{s,k}
=
\underset{1\leq i\leq N_{s,k}}{\arg\min}
\left\|
\mathbf{p}^{\mathrm{query}}_s
-
\Psi(\mathbf{p}_{s,k,i})
\right\|_2,
\qquad
\tilde{\mathbf{p}}_{s,k}
=
\mathbf{p}_{s,k,i^{*}_{s,k}}.
\label{eq:candidate_retrieval}
\end{equation}
The transformed contour points determine the retrieval index. The selected point $\tilde{\mathbf{p}}_{s,k}$ is taken from the rigidly aligned contour, preserving the patient-specific 3D position. Each slice contributes one candidate for the current query. Let $d_{s,k}=\|\mathbf{p}^{\mathrm{query}}_s-\tilde{\mathbf{p}}_{s,k}\|_2$ denote the query-to-candidate distance in the rigidly aligned coordinates. The candidates from the $K$ available SAX slices are aggregated using distance-dependent soft weights:
\begin{equation}
\begin{gathered}
w_{s,k}
=
\frac{
\exp\left[-d_{s,k}^{2}/(2\beta^{2})\right]
}{
\sum_{k'=1}^{K}
\exp\left[-d_{s,k'}^{2}/(2\beta^{2})\right]
},\\
\mathbf{X}^{\mathrm{obs}}_s(u,v)
=
\sum_{k=1}^{K}
w_{s,k}\tilde{\mathbf{p}}_{s,k}.
\end{gathered}
\label{eq:observation_encoding}
\end{equation}
where $\beta$ controls the aggregation bandwidth. Applying this procedure over the canonical UV domain yields the epicardial and endocardial observation fields $\mathbf{X}^{\mathrm{obs}}_{\mathrm{epi}}$ and $\mathbf{X}^{\mathrm{obs}}_{\mathrm{endo}}$.

\subsubsection{Coverage-Aware Slice Sampling}

The number and longitudinal distribution of usable SAX slices vary across acquisitions. To reduce dependence on a fixed slice configuration, we perform coverage-aware sampling on the physical contours before UV encoding.
Let $\mathcal{V}$ denote the ordered set of valid SAX slices. During training, a subset $\mathcal{V}_p\subseteq\mathcal{V}$ is sampled according to the coverage patterns illustrated in Fig.~\ref{fig:method:pipeline}, including full coverage, missing basal or apical slices, retained middle regions, and uniform sparse sampling. The corresponding observation field is written as
\begin{equation}
\mathbf{X}^{\mathrm{obs}}_{s,p}
=
\mathcal{E}
\left(
\{\mathcal{C}_{s,k}\}_{k\in\mathcal{V}_p};
\mathbf{X}^{\mathrm{template}}_s
\right),
\label{eq:coverage_sampling}
\end{equation}
where $\mathcal{E}$ denotes the contour-to-UV encoder.
Because sampling is performed on the physical SAX contours before UV encoding, the resulting inputs preserve realistic longitudinal acquisition patterns. This exposes the network to varying slice coverage during training and improves robustness to incomplete observations.

\subsection{Topology- and Distortion-Aware Surface Completion}
\label{sec:C}

Given the two observation fields, myocardial reconstruction is formulated
as dense coordinate-field completion. The network input is obtained by
concatenating the surface-specific observation fields:
\begin{equation}
\mathcal{I}
=
\operatorname{Concat}
\left(
\mathbf{X}_{\mathrm{epi}}^{\mathrm{obs}},
\mathbf{X}_{\mathrm{endo}}^{\mathrm{obs}}
\right)
\in
\mathbb{R}^{6\times H\times W}.
\label{eq:network_input}
\end{equation}
As shown in Fig.~\ref{fig:method:pipeline}, a shared encoder
$E_{\theta_{\mathrm{e}}}$ extracts a joint representation of the
myocardial geometry, which is subsequently processed by two
surface-specific decoders:
\begin{equation}
\mathbf{Z}
=
E_{\theta_{\mathrm{e}}}(\mathcal{I}),
\qquad
\hat{\mathbf{X}}_{s}
=
D_{\theta_{s}}(\mathbf{Z}),
\quad
s\in\{\mathrm{epi},\mathrm{endo}\}.
\label{eq:surface_completion}
\end{equation}
The shared encoder captures the overall ventricular geometry and the
relationship between the two myocardial boundaries, while each decoder
reconstructs the complete coordinate field of its corresponding surface.

\subsubsection{Circumferentially Periodic Feature Learning}

The original myocardial surface is closed along the circumferential direction, but cutting it into a rectangular UV domain places anatomically adjacent locations at opposite lateral boundaries. Conventional convolution therefore breaks their intrinsic neighborhood.
To preserve this topology, we introduce Circumferential Wrap Convolution:
\begin{equation}
\operatorname{WrapConv2D}(\mathbf{F};\mathbf{G})
=
\mathbf{G}*
\operatorname{Pad}_{v}
\left(
\operatorname{Pad}^{\mathrm{wrap}}_{u}(\mathbf{F})
\right),
\label{eq:wrapconv}
\end{equation}
where $\mathbf{F}$ and $\mathbf{G}$ denote the input feature map and learnable convolution kernel, respectively; periodic padding is applied along the circumferential coordinate $u$, and conventional padding along the longitudinal coordinate $v$.
WrapConv2D is used throughout the encoder and decoders, enabling multi-scale feature aggregation across the UV seam and maintaining circumferential continuity in the completed surface fields.

\subsubsection{Region-Focused Geometric Learning}

The predicted coordinate fields are first supervised by a surface-wide objective:
\begin{equation}
\mathcal{L}_{\mathrm{rec}}
=
\frac{1}{2}
\sum_{s\in\{\mathrm{epi},\mathrm{endo}\}}
\operatorname{MSE}
\left(
M_s\odot\hat{\mathbf{X}}_s,\,
M_s\odot\mathbf{X}^{*}_s
\right),
\label{eq:reconstruction_loss}
\end{equation}
where $M_s$ denotes the valid UV mask. The base objective combines coordinate reconstruction with surface smoothness and global geometric consistency:
\begin{equation}
\begin{aligned}
\mathcal{L}_{\mathrm{base}}
={}&
\mathcal{L}_{\mathrm{rec}}
+\lambda_{\mathrm{tv}}\mathcal{L}_{\mathrm{tv}}
+\lambda_{\mathrm{overlap}}\mathcal{L}_{\mathrm{overlap}}\\
&+
\lambda_{\mathrm{normal}}^{\mathrm{g}}
\mathcal{L}_{\mathrm{normal}}^{\mathrm{g}}
+
\lambda_{\mathrm{flip}}^{\mathrm{g}}
\mathcal{L}_{\mathrm{flip}}^{\mathrm{g}}.
\end{aligned}
\label{eq:base_loss}
\end{equation}
Here, $\mathcal{L}_{\mathrm{tv}}$ smooths the UV coordinate fields, $\mathcal{L}_{\mathrm{overlap}}$ preserves the coincidence of template-defined shared epicardial-endocardial vertices, and $\mathcal{L}_{\mathrm{normal}}^{\mathrm{g}}$ and $\mathcal{L}_{\mathrm{flip}}^{\mathrm{g}}$ constrain surface-wide normal consistency and face inversion, respectively.
$\lambda_{\mathrm{tv}}$, $\lambda_{\mathrm{overlap}}$, $\lambda^g_{\mathrm{normal}}$, and $\lambda^g_{\mathrm{flip}}$ are all balancing parameters.

Although the canonical UV representation provides a regular domain for structured surface learning, the contraction of the ventricular circumference near the apex still introduces pronounced metric distortion. Coordinate errors in this region are therefore more likely to be amplified into local mesh artifacts. Because these parameterization-sensitive locations occupy only a small portion of the surface, surface-wide averaging may not provide sufficient emphasis. 
We thus precompute a focus core $\mathcal{R}_0$ from canonical-template
locations exhibiting strong metric compression or UV-grid conflicts.
Surrounding buffer regions $\{\mathcal{R}_r\}_{r=1}^{R}$ are constructed
through mesh-adjacency expansion. For each geometric term
$q\in\mathcal{Q}=\{\mathrm{edge},\mathrm{area},\mathrm{normal},\mathrm{flip}\}$,
we compute a region-balanced loss as
\begin{equation}
\mathcal{L}^{\mathrm{focus}}_q
=
\frac{
\sum_{r=0}^{R}\gamma_r\mathcal{L}^{(r)}_q
}{
\sum_{r=0}^{R}\gamma_r
},
\label{eq:regional_loss}
\end{equation}
where $\mathcal{L}^{(r)}_q$ is the mean penalty within $\mathcal{R}_r$
and $\gamma_r$ controls its contribution. This region-wise balancing
prevents the compact focus core from being dominated by larger surrounding
regions. The overall region-focused objective is
\begin{equation}
\mathcal{L}_{\mathrm{focus}}
=
\sum_{q\in\mathcal{Q}}
\lambda_q\mathcal{L}^{\mathrm{focus}}_q,
\label{eq:focus_loss}
\end{equation}
where $\lambda_q$ is balancing parameter.
Therefore, the complete objective is
$\mathcal{L}_{\mathrm{total}}
=\mathcal{L}_{\mathrm{base}}+\mathcal{L}_{\mathrm{focus}}$.
Finally, the predicted epicardial and endocardial coordinate fields are converted into explicit triangular meshes using the fixed connectivity of the canonical UV grid.

\begin{table*}[t]
\centering
\caption{Quantitative comparison of reconstruction on ACDC, M\&Ms, and M\&Ms-2. Results are reported as mean (standard deviation).}
\footnotesize

\begin{tabular}{lccccccc}
\toprule
\multirow{2}{*}{Method} &
\multicolumn{2}{c}{ACDC} &
\multicolumn{2}{c}{M\&Ms} &
\multicolumn{2}{c}{M\&Ms-2} &
\multirow{2}{*}{Inference} \\
\cmidrule(lr){2-3}
\cmidrule(lr){4-5}
\cmidrule(lr){6-7}
&
CD (mm) $\downarrow$ & $F@2\mathrm{mm}$ $\uparrow$
&
CD (mm) $\downarrow$ & $F@2\mathrm{mm}$ $\uparrow$
&
CD (mm) $\downarrow$ & $F@2\mathrm{mm}$ $\uparrow$
&
Time (s/frame) $\downarrow$ \\
\midrule
Voxel2Mesh
& 8.629 (2.773) & 0.230 (0.126)
& 8.325 (2.742) & 0.246 (0.120)
& 8.560 (2.761) & 0.246 (0.125)
& 1.4 \\
PN-GCN
& 3.183 (0.395) & 0.741 (0.092)
& 2.873 (0.446) & 0.808 (0.092)
& 3.144 (0.838) & 0.770 (0.106)
& 1.0 \\
4DMM
& 2.983 (0.398) & 0.782 (0.067)
& 2.776 (0.339) & 0.820 (0.064)
& 2.914 (0.387) & 0.797 (0.066)
& $>30$ \\
GHDHeart
& 4.700 (1.060) & 0.463 (0.112)
& 4.528 (1.169) & 0.511 (0.131)
& 4.684 (1.012) & 0.479 (0.117)
& $>30$ \\
\textbf{Ours}
& \textbf{2.887 (0.388)} & \textbf{0.796 (0.083)}
& \textbf{2.641 (0.381)} & \textbf{0.839 (0.076)}
& \textbf{2.810 (0.524)} & \textbf{0.817 (0.076)}
& \textbf{$0.8$} \\
\bottomrule
\end{tabular}
\label{compare_tab_3d_sax}
\end{table*}

\begin{table*}[t]
\centering
\caption{Quantitative comparison of left ventricular (LV) myocardium segmentation obtained by intersecting the reconstructed surfaces with the original SAX planes. Best and second-best results are shown in bold and underlined, respectively.}
\label{tab:segmentation_comparison}
\scriptsize
\begin{tabular}{lccccccccc}
\toprule
\multirow{2}{*}{Method}
& \multicolumn{3}{c}{ACDC}
& \multicolumn{3}{c}{M$\&$Ms}
& \multicolumn{3}{c}{M$\&$Ms-2} \\
\cmidrule(lr){2-4}
\cmidrule(lr){5-7}
\cmidrule(lr){8-10}
& Dice $\uparrow$
& ASSD (mm) $\downarrow$
& HD95 (mm) $\downarrow$
& Dice $\uparrow$
& ASSD (mm) $\downarrow$
& HD95 (mm) $\downarrow$
& Dice $\uparrow$
& ASSD (mm) $\downarrow$
& HD95 (mm) $\downarrow$ \\
\midrule
Voxel2Mesh
& 0.506 (0.200)
& 2.237 (1.513)
& 9.171 (5.059)
& 0.531 (0.162)
& 2.082 (1.345)
& 9.615 (4.185)
& 0.481 (0.132)
& 3.468 (1.245)
& 17.586 (4.575) \\
PN-GCN
& 0.899 (0.044)
& 0.329 (0.295)
& 2.616 (3.368)
& 0.923 (0.043)
& 0.280 (0.394)
& \underline{2.664 (3.733)}
& 0.893 (0.070)
& 0.398 (0.668)
& 3.409 (5.186) \\
4DMM
& \textbf{0.956 (0.024)}
& \textbf{0.172 (0.205)}
& \textbf{2.386 (3.011)}
& \underline{0.953 (0.028)}
& \underline{0.227 (0.316)}
& 2.765 (3.574)
& \underline{0.944 (0.031)}
& \underline{0.250 (0.323)}
& \underline{3.203 (4.014)} \\
GHDHeart
& 0.806 (0.071)
& 0.790 (0.489)
& 4.448 (2.898)
& 0.786 (0.062)
& 1.216 (0.720)
& 8.979 (5.262)
& 0.750 (0.069)
& 1.256 (0.684)
& 9.597 (5.397) \\
\textbf{Ours}
& \underline{0.939 (0.031)}
& \underline{0.212 (0.213)}
& \underline{2.429 (2.654)}
& \textbf{0.965 (0.021)}
& \textbf{0.114 (0.204)}
& \textbf{1.181 (2.327)}
& \textbf{0.949 (0.045)}
& \textbf{0.193 (0.512)}
& \textbf{1.932 (4.029)} \\
\bottomrule
\end{tabular}
\end{table*}

\section{Experiments and Results}

\subsection{Datasets and Experimental Protocols}

We evaluated the reconstruction framework on three public cine MRI datasets: ACDC~\citep{bernard2018deep}, M$\&$Ms~\citep{campello2021multi}, and M$\&$Ms-2~\citep{martin2023deep}. ACDC contains SAX cine MRI from 150 cases evenly distributed into five diagnostic categories, including normal subjects (NOR), dilated cardiomyopathy (DCM), hypertrophic cardiomyopathy (HCM), myocardial infarction with altered left ventricular ejection fraction (MINF), and abnormal right ventricle (ARV). 
From M\&Ms and M\&Ms-2, we included 340 and 350 cases, respectively. 
These datasets provide complementary multi-center, multi-vendor, and multi-disease cohorts for evaluating reconstruction under heterogeneous acquisition conditions and cardiac anatomies.
For reconstruction, each dataset was independently split at the subject level into approximately $70\%$ training and $30\%$ testing, with $20\%$ of the training subjects reserved for validation. The test sets remained unseen until final evaluation.

For downstream cine MRI-based scar localization (Sec.~\ref{exp:application:scar}), we additionally used the CineMyoPS dataset from the CARE challenge~\citep{ding2025cinemyops}, comprising 64 cases from two centers with cine MRI and manual annotations of the LV, myocardium, and myocardial scar at end diastole. The trained reconstruction model was used to generate UV-domain myocardial surface sequences from cine SAX inputs, from which abnormal motion features were subsequently used for scar prediction.
For UV-domain generative modeling, the reconstructed UV sequences from 270 NOR, DCM, and HCM cases were used to learn disease-conditioned myocardial motion generation, with 216 cases for training and 54 held-out cases for evaluation.

\subsection{Reference Mesh Construction and Evaluation}

All methods used SAX contours generated by a fixed pretrained nnU-Net~\citep{isensee2021nnu}. Patient-specific reference meshes with consistent topology and vertex correspondence were generated offline from the multi-phase SAX contours using a public cardiac atlas~\citep{bai2015bi} through global alignment, non-rigid refinement, and temporal smoothing.
Surface agreement with these references was evaluated using symmetric Chamfer distance (CD) and F-score at a 2-mm tolerance (F@2mm). SAX-plane agreement was assessed against the corresponding segmentations using Dice, ASSD, and HD95 after intersecting the reconstructed surfaces with the SAX planes. Cross-phase motion agreement was evaluated using ED-relative motion endpoint error (EPE)~\citep{qiao2020temporally} at corresponding UV locations.
Functional agreement was evaluated using end-diastolic volume (EDV) and left ventricular ejection fraction (LVEF).
For the ablation studies, we additionally report UV MSE, measuring coordinate-field reconstruction error over the valid UV domain, and Edge Ratio P95, measuring high-percentile local mesh distortion relative to the reference surface. 

\subsection{Implementation}

All experiments were implemented in PyTorch and conducted on a single NVIDIA GeForce RTX 4090 GPU. 
All coordinate fields were represented on a $256\times256$ canonical UV grid, with an aggregation bandwidth of $\beta=0.06$. 
The concatenated epicardial and endocardial fields formed a six-channel input to a seam-aware 2D U-Net with a shared encoder and surface-specific decoders. 
Separate models were trained from scratch for ACDC, M\&Ms, and M\&Ms-2 using full- and incomplete-coverage samples generated by coverage-aware sampling. 
AdamW was used with a batch size of $16$, learning rate $10^{-4}$, weight decay $10^{-2}$, gradient clipping at $1.0$, and cosine annealing; training lasted $48{,}000$, $75{,}000$, and $80{,}000$ steps, respectively. 
The surface-wide and region-focused loss weights were set to $(0.01,0.1,0.003,0.001)$ and $(0.04,0.005,0.003,0.002)$, respectively, with region weights $(1.00,0.75,0.50,0.25)$.
Checkpoints were selected using the validation set only.

\begin{figure*}[!t]
\includegraphics[width=0.97\textwidth]{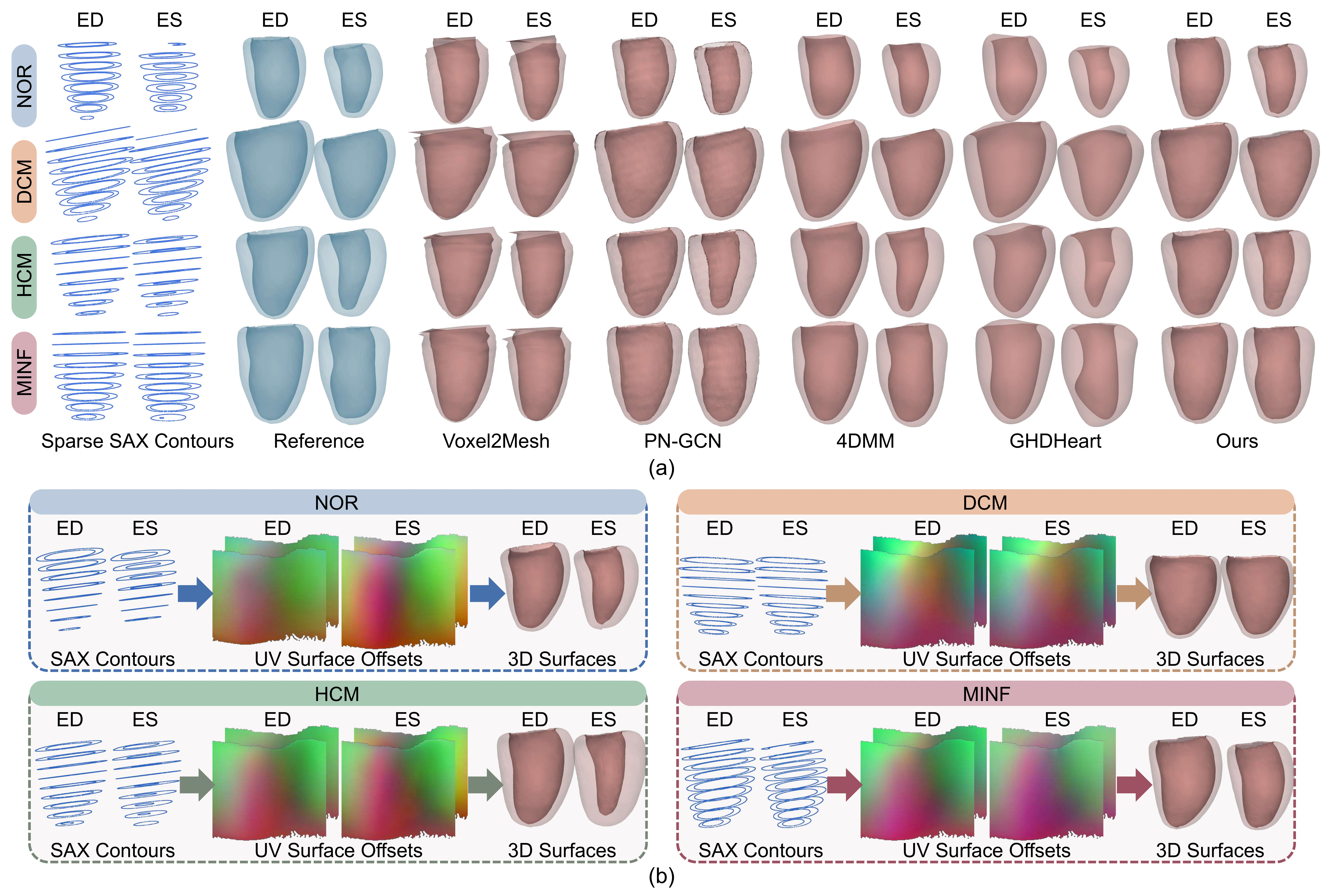}
\centering
\vspace{-10pt}
\caption{Qualitative comparison of disease-specific myocardial reconstruction:
(a) ED and ES surfaces for representative NOR, DCM, HCM, and MINF cases, showing sparse SAX contours, reference surfaces, and predictions from the competing methods. 
(b) Corresponding SAX contours, UV surface-offset maps, and reconstructed 3D surfaces. For visualization, UV fields show normalized template-relative 3D offsets in RGB. NOR: normal subjects; DCM/HCM: dilated/hypertrophic cardiomyopathy; MINF: myocardial infarction with altered LV ejection fraction.
}
\label{fig:compare_3d_sax}
\end{figure*}


\begin{figure}[!t]
\includegraphics[width=1\columnwidth]{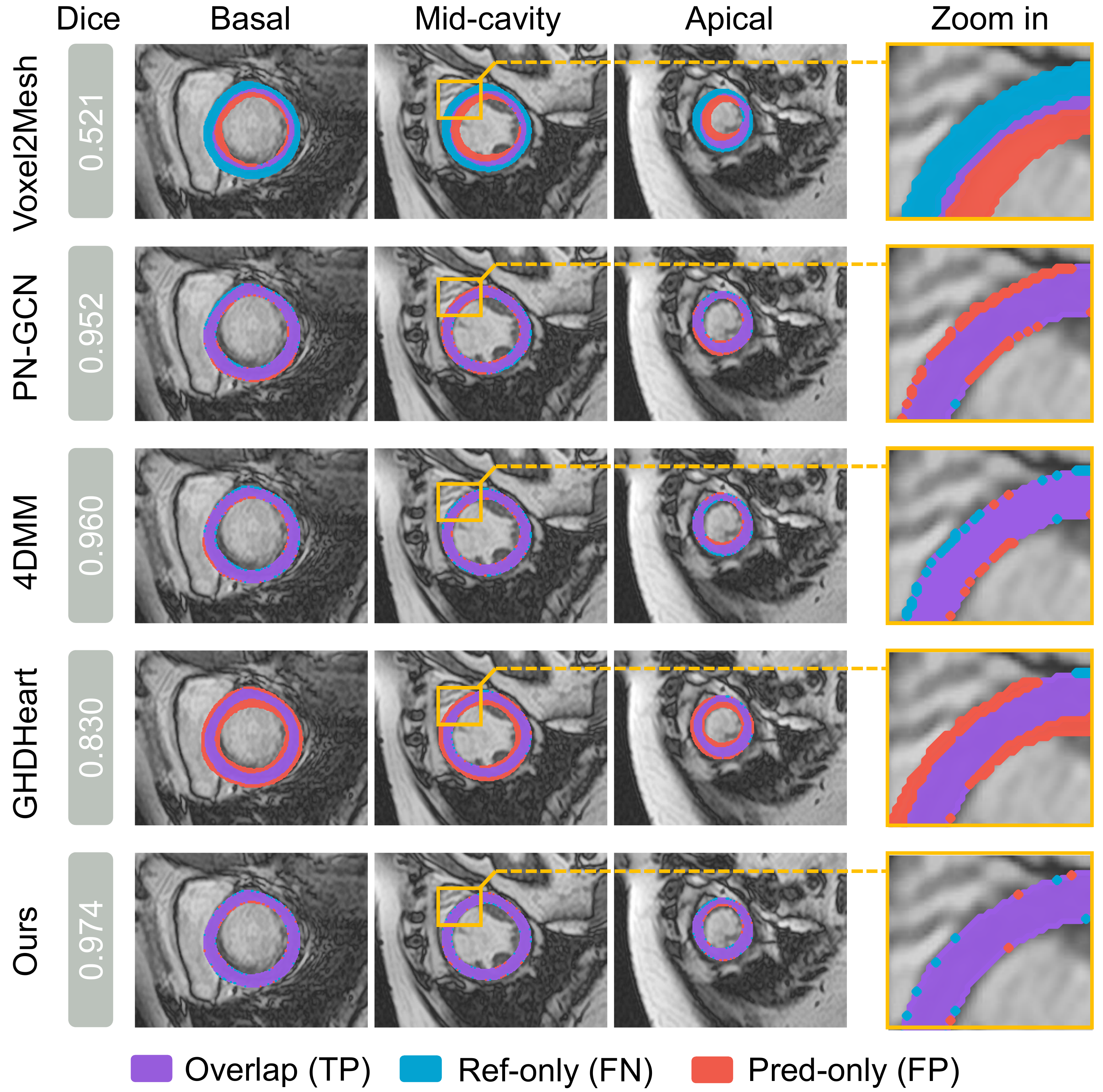}
\centering
\vspace{-7pt}
\caption{Qualitative comparison of LV myocardial segmentation obtained by intersecting the reconstructed surfaces with the original SAX planes. }
\label{fig:compare_seg_sax2}
\end{figure}

\subsection{Comparison Study}

We compared with four representative approaches reproducible under the same sparse-SAX setting: Voxel2Mesh~\citep{wickramasinghe2020voxel2mesh}, using volumetric features for mesh deformation; a PN-GCN baseline implemented in this study, encoding the same contours as unordered point features without explicit UV correspondence; 4DMM~\citep{conf/ICCV/yuan2023}, using implicit shape and motion fields; and GHDHeart~\citep{journal/MedIA/luo2026}, combining differentiable slicing with global mesh deformation. All trainable baselines were retrained using the same subject-level splits. As shown in Table~\ref{compare_tab_3d_sax}, our method achieved the best surface reconstruction accuracy on all three datasets.
The CD values were $2.887$~mm on ACDC, $2.641$~mm on M$\&$Ms, and $2.810$~mm on M$\&$Ms-2, with corresponding F@2mm values of $0.796$, $0.839$, and $0.817$.
Our method significantly outperformed the best baseline on both metrics across all datasets (Holm-corrected paired Wilcoxon tests, $p\leq0.005$).
The qualitative results in Fig.~\ref{fig:compare_3d_sax} (a) further demonstrated stable recovery of disease-specific ED and ES morphologies. 
Compared with these approaches, the proposed contour-to-UV encoding explicitly associates sparse SAX observations with anatomically corresponding surface locations before completion, allowing local contour evidence to directly guide structured surface prediction.

Table~\ref{tab:segmentation_comparison} further evaluated the LVM segmentations obtained by intersecting the reconstructed surfaces with the original SAX planes.
On ACDC, our method performed comparably to 4DMM, achieving a Dice of $0.939$, an ASSD of $0.212$~mm, and an HD95 of $2.429$~mm. 
On the more heterogeneous M\&Ms and M\&Ms-2 datasets, it achieved the best results across all metrics, with Dice scores of $0.965$ and $0.949$, ASSD values of $0.114$ and $0.193$~mm, and HD95 values of $1.181$ and $1.932$~mm, respectively. 
These improvements were consistent with the basal, mid-cavity, and apical comparisons shown in Fig.~\ref{fig:compare_seg_sax2}. 
Moreover, the proposed method required $0.8$~s per frame, compared with more than $30$~s for 4DMM and GHDHeart. 
Fig.~\ref{fig:compare_3d_sax} (b) further demonstrated that patient- and phase-specific variations were represented within a shared circumferential-longitudinal UV domain, enabling explicit anatomical correspondence and efficient reconstruction. 
The Supporting Video visualizes disease-specific reconstruction, method comparisons, and robustness to varying slice coverage.

\begin{table}[t]
\centering
\caption{Ablation study of the proposed framework on ACDC. UV MSE is reported in units of~$(\times10^{-3})$.}
\scriptsize
\begin{tabular}{lccc}
\toprule
Variant
& UV MSE $\downarrow$
& CD (mm) $\downarrow$
& Edge Ratio P95 $\downarrow$ \\
\midrule
Rigid / Full-only
& 1.565 (3.413)
& 3.569 (1.829)
& 1.447 (0.187) \\
Centerline / Full-only
& 1.491 (3.106)
& 3.538 (1.718)
& 1.449 (0.189) \\
w/o Region Focus
& 0.499 (0.541)
& 3.021 (0.571)
& 1.408 (0.053) \\
w/o WrapConv2D
& 0.535 (0.573)
& 2.999 (0.532)
& 1.406 (0.052) \\
Ours
& \textbf{0.487 (0.531)}
& \textbf{2.997 (0.546)}
& \textbf{1.393 (0.049)} \\
\bottomrule
\end{tabular}
\label{tab:ablation1}
\end{table}

\begin{table}[t]
\centering
\caption{Reconstruction accuracy under five slice-coverage patterns for models trained with full coverage only and with coverage-aware sampling.}
\scriptsize
\begin{tabular}{lcccc}
\toprule
& \multicolumn{2}{c}{UV MSE $\downarrow$}
& \multicolumn{2}{c}{CD (mm) $\downarrow$} \\
\cmidrule(lr){2-3}
\cmidrule(lr){4-5}
Pattern
& Full-only
& Coverage-aware
& Full-only
& Coverage-aware \\
\midrule
Full
& \textbf{0.383} & 0.408
& \textbf{2.852} & 2.888 \\
Base-missing
& 1.380 & \textbf{0.544}
& 3.913 & \textbf{3.089} \\
Apex-missing
& 0.998 & \textbf{0.520}
& 3.303 & \textbf{2.989} \\
Mid-block
& 10.868 & \textbf{0.963}
& 8.464 & \textbf{3.702} \\
Uniform-sparse
& 2.122 & \textbf{0.598}
& 4.302 & \textbf{3.139} \\
\bottomrule
\end{tabular}
\label{tab:ablation2}
\end{table}

\begin{figure*}[!t]
\center
\includegraphics[width=0.84\linewidth]{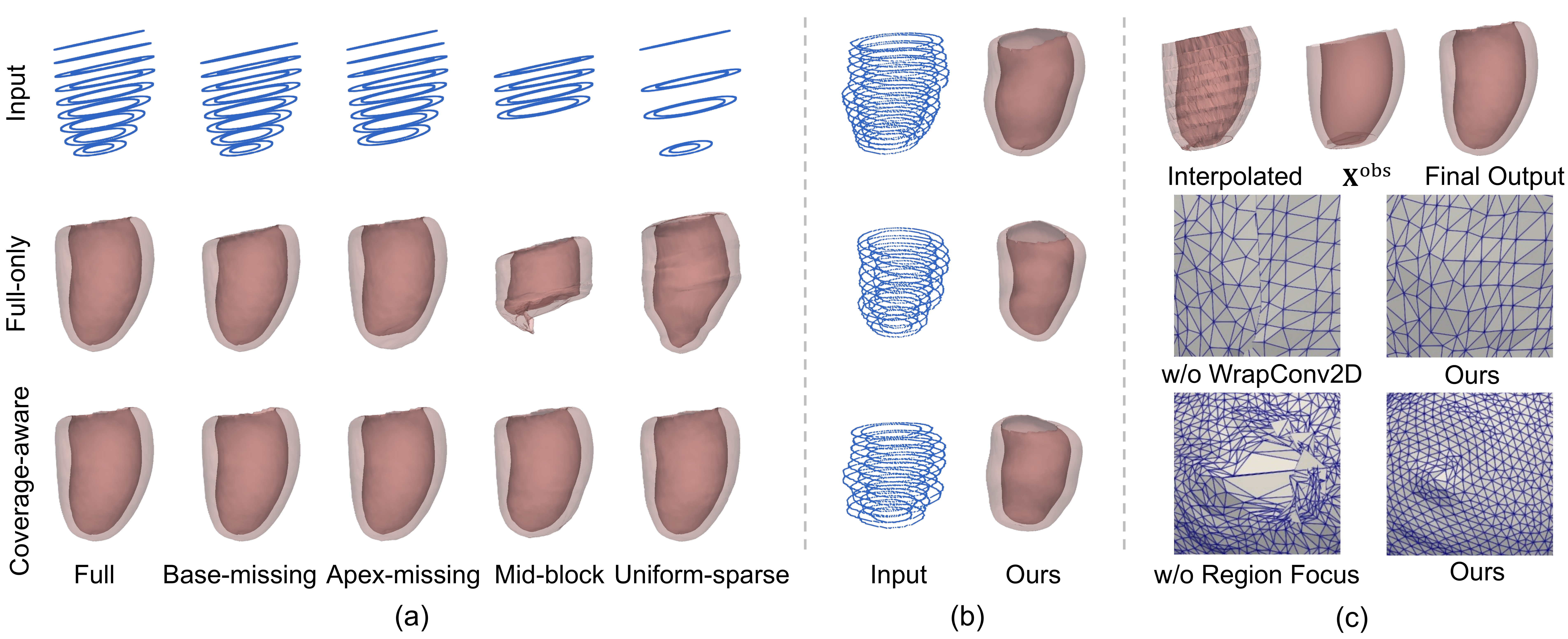}
\caption{
(a) Full-only and coverage-aware reconstructions under five slice-coverage patterns.
(b) Representative results for locally irregular contour observations.
(c) Comparison of non-learning UV reconstruction and network completion (top), seam continuity without and with WrapConv2D (middle), and apical geometry without and with region-focused supervision (bottom).}
\label{fig:abla}
\end{figure*}

\subsection{Ablation Study}

Tables~\ref{tab:ablation1} and~\ref{tab:ablation2} evaluated the contributions of centerline alignment, coverage-aware training, region-focused supervision, and WrapConv2D. 
The \textit{Rigid / Full-only} and \textit{Centerline / Full-only} variants were trained only with full-coverage samples, whereas the complete model incorporated all proposed components. 
Results were pooled over five coverage patterns and jointly computed for the epicardial and endocardial surfaces using UV MSE, CD, and Edge Ratio P95, where values closer to $1$ indicated lower local edge distortion. 
The complete model achieved the best performance across all three metrics. 
Compared with \textit{Centerline / Full-only}, it reduced UV MSE and CD by $67.3\%$ and $15.3\%$, respectively. 
Table~\ref{tab:ablation2} and Fig.~\ref{fig:abla} (a) showed that coverage-aware training caused only a minor change under full coverage but consistently improved all incomplete-coverage settings. 
The largest gain occurred for the mid-block pattern, where UV MSE and CD were reduced by $91.1\%$ and $56.3\%$, respectively, demonstrating improved robustness to base-missing, apex-missing, mid-block, and uniformly sparse SAX inputs.

The similar performance of the two full-only variants suggested that both rigid and centerline alignment supported reliable reconstruction under complete coverage. Centerline alignment was nevertheless retained to provide a consistent geometric basis for correspondence retrieval across slices and subjects. 
As shown in Fig.~\ref{fig:abla} (b), the learned UV-domain completion model also preserved globally smooth and coherent surfaces in the presence of local contour misalignment or irregular variation. 
Fig.~\ref{fig:abla} (c) further showed that direct UV interpolation produced stripe-like artifacts and that direct conversion of $\mathbf{X}^{\mathrm{obs}}$ recovered only locally observed geometry, whereas the proposed network generated complete and smooth myocardial surfaces. 
WrapConv2D improved continuity across the circumferential seam, while region-focused supervision reduced triangle stretching and distortion near the apex. These observations were consistent with the improvements in UV MSE and Edge Ratio P95 and indicated that global CD alone did not fully reflect localized geometric defects.

\begin{figure*}[ht]
    \centering
    \includegraphics[width=1.0\linewidth]{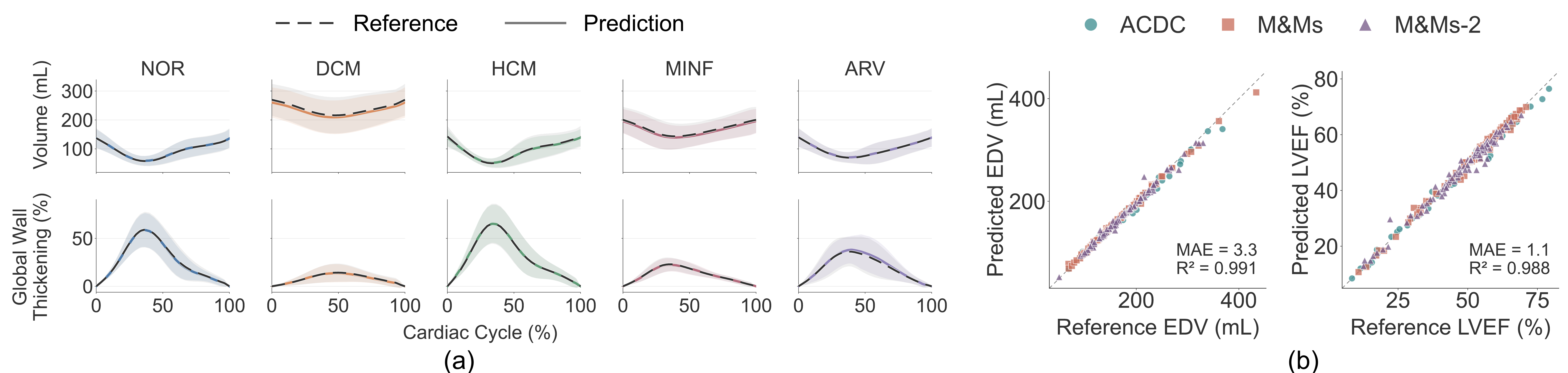}
    \vspace{-15pt}
    \caption{Quantitative analysis of myocardial shape and function. (a) Predicted and reference LV volume and global wall-thickening trajectories for the five ACDC diagnostic groups; shaded regions indicate group-wise standard deviations. (b) Agreement between predicted and reference EDV and LVEF across ACDC, M\&Ms, and M\&Ms-2. ARV: abnormal right ventricle.}
    \label{fig:shape_motion_analysis}
\end{figure*}

\begin{figure}[t]
    \centering
    \includegraphics[width=1.0\linewidth]{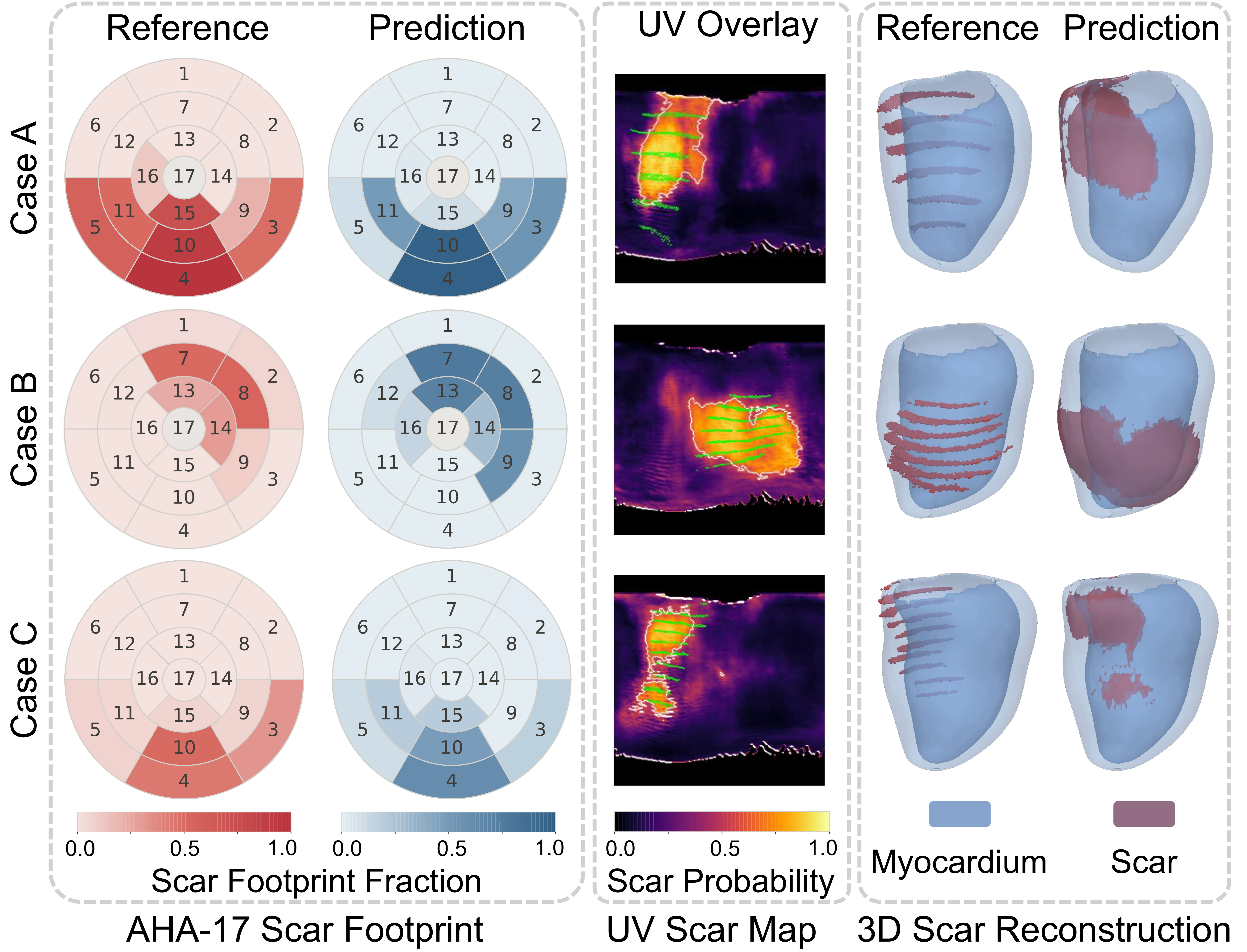}
    \vspace{-7pt}
    \caption{Qualitative myocardial scar localization in three representative CineMyoPS cases.}
    \label{fig:scar-pred}
\end{figure}

\subsection{Utility of the Canonical UV Representation}

\subsubsection{Myocardial Shape and Functional Quantification}

The reconstructed surfaces were used to derive frame-wise LV volume and global wall-thickening trajectories, with wall thickening computed from temporal changes in the paired epicardial-endocardial distances on the shared UV domain. 
As shown in Fig.~\ref{fig:shape_motion_analysis} (a), the predicted curves closely followed the references throughout the cardiac cycle and preserved disease-specific patterns, including enlarged ventricular volumes and reduced wall thickening in DCM and MINF, and increased thickening in HCM. Across the three datasets, the case-level motion EPE was $1.297\pm0.643$~mm, indicating accurate point-wise motion across cardiac phases.
Fig.~\ref{fig:shape_motion_analysis} (b) further showed strong agreement in functional measurements across the three datasets, with an EDV MAE of $3.3$~mL ($R^2=0.991$) and an LVEF MAE of $1.1\%$ ($R^2=0.988$). 
These results demonstrated that the reconstructed 4D surfaces preserved both disease-related myocardial dynamics and clinically relevant ventricular function.

\subsubsection{UV-Domain Myocardial Scar Localization from Cine MRI} \label{exp:application:scar}

To further investigate whether the reconstructed 4D myocardial geometry captured clinically relevant regional motion abnormalities, we evaluated myocardial scar footprint localization on the independent CineMyoPS dataset~\citep{ding2025cinemyops}. 
Here, the scar footprint referred to the surface region obtained by projecting the myocardial scar along the wall onto the corresponding myocardial surface. From the reconstructed epicardial and endocardial sequences, we derived motion amplitude, fractional wall thickening, and cumulative displacement over the cardiac cycle. 
These features were compared with a healthy motion atlas constructed from approximately 200 normal cases to identify anatomically localized reductions in contraction. 
Because the reconstructed surfaces maintained point-wise correspondence across cardiac phases and subjects, the resulting abnormal-motion patterns could be directly mapped to LGE-derived scar regions in the shared UV domain. 
A lightweight predictor with WrapConv2D subsequently converted these motion-abnormality maps into continuous scar-footprint probability maps. 
As shown in Fig.~\ref{fig:scar-pred}, the predicted scar footprints were spatially consistent with the references in the AHA-17 representation, UV domain, and reconstructed 3D myocardium, indicating that the reconstructed 4D geometry preserved regional motion characteristics relevant to myocardial scar localization.

\begin{figure}[t]
    \centering
    \includegraphics[width=1\linewidth]{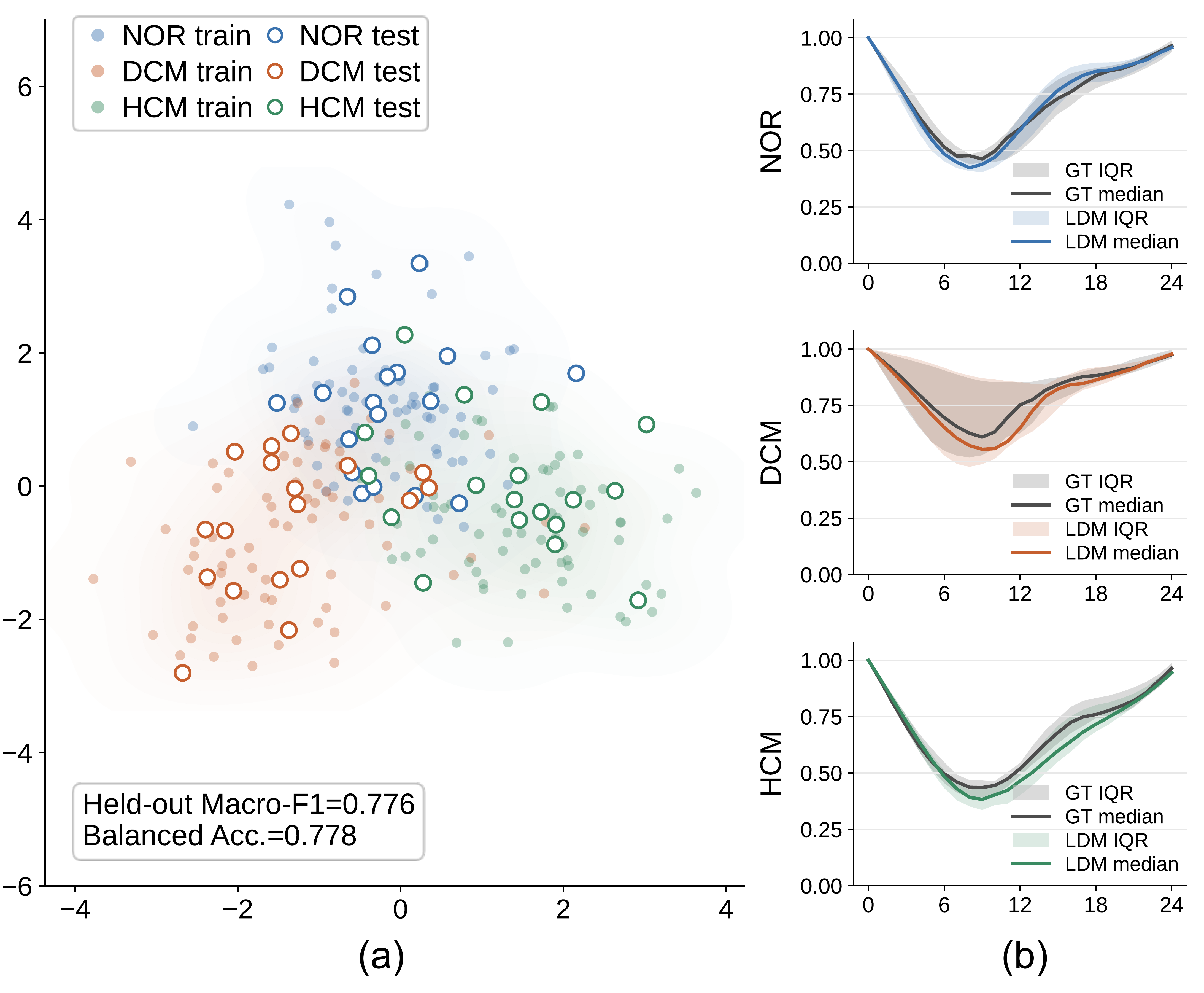}
    \vspace{-7pt} %
    \caption{Conditional myocardial motion generation and analysis in the UV domain. (a) Projection of the VAE motion latents onto the first two linear discriminant axes, showing disease-related distributions of NOR, DCM, and HCM. (b) Reference and generated normalized LV cavity-volume curves, where the horizontal axis denotes the normalized cardiac phase and the vertical axis denotes $V(t)/V(0)$. IQR: interquartile range.}
    \label{fig:generation}
\end{figure}



\subsubsection{UV-Domain Generative Modeling}

To demonstrate the broader utility of the canonical UV representation, temporally aligned epicardial and endocardial motion was converted into regular UV displacement fields and modeled using a sequence VAE followed by a conditional latent diffusion model (LDM). Each UV sequence was first decomposed into the ED UV anatomy and 24 frames of ED-relative motion. The VAE compressed each $24\times6\times256\times256$ motion sequence into an $8\times32\times32$ latent code. Conditioned on the ED UV and disease label, the LDM then generated a motion latent, which was decoded by the frozen VAE decoder and added back to the ED anatomy to recover a complete 25-frame UV sequence.
Although the VAE was optimized solely for motion reconstruction, its latent space retained clear disease-related structure. As shown in Fig.~\ref{fig:generation} (a), NOR, DCM, and HCM exhibit distinguishable distributions along the first two linear discriminant axes. A classification probe trained on the motion latents achieved a held-out macro-F1 of $0.776$ and a balanced accuracy of $0.778$, indicating that the shared UV correspondence organizes temporal motion into a compact representation with consistent spatial semantics while preserving disease-related motion characteristics.
The LDM generated complete 25-frame motion sequences for all held-out cases, achieving an overall CD of $3.929$~mm. Fig.~\ref{fig:generation} (b) further shows that the generated sequences preserve the characteristic LV cavity-volume patterns of NOR, DCM, and HCM and remain consistent with the reference distributions. These results demonstrate that the canonical UV representation not only provides consistent anatomical correspondence across subjects and cardiac phases, but also enables conventional 2D generative models to capture disease-related 4D myocardial motion while retaining geometric and functional fidelity after reconstruction in 3D.

\section{Discussion and Conclusion} 

This study introduced an anatomy-aligned UV representation for patient-specific 4D myocardial reconstruction from sparse SAX cine MRI. By unfolding the epicardial and endocardial surfaces onto a shared circumferential-longitudinal domain, the framework transformed the irregular 3D reconstruction problem into structured two-dimensional coordinate-field completion and directly associated sparse observations with anatomically corresponding surface locations. This shared organization preserved explicit anatomical correspondence across subjects and cardiac phases.
The method achieved the best surface accuracy across ACDC, M\&Ms, and M\&Ms-2 with an inference time of only $0.8$~s per frame, and showed particularly strong performance on the more heterogeneous M\&Ms cohorts. The close agreement in EDV and LVEF, with MAEs of $3.3$~mL and $1.1\%$, respectively, further indicated that the reconstructed sequences preserved clinically relevant ventricular geometry and function.

The experimental results clarified the contributions of the individual components. Coverage-aware sampling provided the largest improvement by reducing dependence on a fixed longitudinal slice distribution; compared with full-coverage training, it substantially improved reconstruction under missing basal, apical, mid-block, and uniformly sparse observations. WrapConv2D maintained continuity across the circumferential seam, whereas region-focused supervision reduced localized distortion near the apex, demonstrating that global surface-distance metrics alone were insufficient to characterize geometric quality in parameterization-sensitive regions. Beyond reconstruction, the shared UV correspondence supported several complementary applications. The reconstructed 4D geometry captured regional reductions in motion and wall thickening that could be mapped to LGE-derived scar regions in the independent CineMyoPS cohort. The same representation also converted myocardial motion into regular UV displacement fields compatible with conventional VAE and latent diffusion architectures. The resulting latent features retained disease-related motion information, while the generated trajectories preserved both surface geometry and global ventricular function. Together, these experiments showed that the canonical UV domain served not only as a reconstruction space but also as a common representation for functional analysis, pathology localization, and generative modeling.

Several limitations remain. 
Separate models were trained for each dataset, and the current formulation focused on the left-ventricular myocardium; extending the framework to a unified multi-dataset model and more complex whole-heart anatomy will require additional parameterization strategies. Although periodic convolution and region-focused constraints reduced seam and apical artifacts, reconstruction quality still depended on the canonical template and its UV mapping. The scar-localization experiment was primarily qualitative and used LGE-derived labels for supervision, while the generative evaluation involved a relatively small cohort containing only NOR, DCM, and HCM cases. 
Future work should employ larger external cohorts and include quantitative scar assessment, uncertainty estimation, and evaluation in downstream simulation and prognostic tasks. Despite these limitations, anatomy-aligned UV surface-field learning provided an accurate, efficient, and extensible framework for correspondence-aware 4D myocardial reconstruction from sparse cine MRI.

\bibliographystyle{model2-names}
\biboptions{authoryear}
\bibliography{A_refs_full}

\end{document}